# Region Based Extensive Response Index Pattern for Facial Expression Recognition

Monu Verma, S. K. Vipparthi, Girdhari Singh
Dept. of Computer Science and Engineering.
Malviya National Institute of Technology, Jaipur, India
monuverma1010@gmail.com, skvipparthi@mnit.ac.in, gsingh.cse@mnit.ac.in

*Abstract*— **This paper presents a novel descriptor named Region based Extensive Response Index Pattern (RETRaIN) for facial expression recognition. The RETRaIN encodes the relation among the reference and neighboring pixels of facial active regions. These relations are computed by using directional compass mask on an input image and extract the high edge responses in foremost directions. Further extreme edge index positions are selected and encoded into six-bit compact code to reduce feature dimensionality and distinguish between the uniform and non-uniform patterns in the facial features. The performance of the proposed descriptor is tested and evaluated on three benchmark datasets Extended Cohn Kanade, JAFFE, and MUG. The RETRaIN achieves superior recognition accuracy in comparison to state-of-the-art techniques.**

## I. INTRODUCTION

Expression appearance on face plays an important role in any human interaction as they convey information about mental status and intention of a person during the communication. Therefore, designing an algorithm for automatic facial emotion recognition becomes a challenging task in computer vision and learning. Many demanding applications are there in these fields, for example: automatic reading tutor analyzes the emotions of a student to improve e-learning system; for medical diagnosis a doctor could analyze the behavior of patients and conclude accordingly; in online shopping, insight into interest of purchaser might lead to improvement in sales. Facial expressions were first investigated by Darwin [1] in 1872. After a long time, 1960s psychologist Paul Ekman [2] came up with the concept of six primitive emotion states (Anger, Sadness, Happiness, Disgust, Surprise, and Fear) called as universal expressions. These emotional states were considered as a standard for Facial Expression Recognition (FER) systems. In 1977 Paul Ekman and Friense created a Facial Action Coding [3] system, which provides information about the active regions (eyes corners, eyebrows, lips corner, nose, and mouth) of the human face. The key challenge of FER algorithms is extracting more informatics features from these active regions. On the basis of feature extraction techniques, FER framework can be partitioned into two classes: Geometry and Appearance based systems. Geometry based feature extraction techniques extract the information by using shapes, angle, and distances between Action Coding (AC) facial points. On the other hand,

Appearance based feature techniques depend on the pixel's intensity value to extract texture information from completes image or regions within image. Usually, in appearance-based techniques local/global filters are used to analyze the appearance/pattern of the image. Numerous Geometric and appearance based methods are available in literature, Few from each category are mentioned here; geometry based methods: Fisher Independent Component Analysis (FICA) [4], Angular descriptor based on Conditional Random Fields (CRF) [5], Voronoi diagram based [6], and graph based [7] [8] etc., appearance based methods: fisher faces [9], Eigen faces [10], Local Discriminate Analysis (LDA) [11], Gabor feature [12], Histogram of Gradients (HOG) [13], and Local Binary Pattern (LBP) [14].Out of these LBP gives highly accurate results; making it one of the most popular approach. But, LBP suffers from white noise and illumination variation problem. To improve the power of LBP many extended approaches are investigated like Local Ternary Pattern (LTP) [15], Local Directional Pattern (LDP) [16], Local directional Number (LDN) [17], and Local Directional Ternary Pattern (LDTP) [18], etc. Recent methods used the directional information in place of the actual intensity value of the neighbors. These approaches show comparatively better results, but still; these techniques have common issues arising due variation in age, pose, and illumination etc. Although, new approaches have been introduced in literature like gradient faces [19], to cater to the issue of variation, but resulting in lower accuracy.

Taking inspiration from directional pattern descriptors, we propose a novel descriptor RETRaIN: A robust technique for emotion recognition using facial expressions under variable conditions. In this paper, directional information is extracted in place of actual intensity values, to overcome the issues of white noise and non-monotonic illumination variations. To achieve this, a compass mask [22] is applied followed by computation of edge responses of input image in eight directions. This compass mask has ability to extract the edge responses of highly affected emotion related features. After this extraction of two max response index values as the directional number from referencing neighborhood and extensive neighborhood is done respectively. These directions provide the structural information of the face. Analysis of the structural information is done to differentiate edges from smooth areas.

Proposed descriptor utilizes the direction information to encode the texture of face. For that, edge responses are computed in the local and extensive neighborhood by applying compass mask in eight directions. Then, two prominent

$$\begin{bmatrix} -1 & -1 & 2 \\ -1 & -1 & 2 \\ -1 & -1 & 2 \end{bmatrix} \begin{bmatrix} -1 & 2 & 2 \\ -1 & -1 & 2 \\ -1 & -1 & 1 \end{bmatrix} \begin{bmatrix} 2 & 2 & 2 \\ -1 & -1 & -1 \\ -1 & -1 & -1 \end{bmatrix} \begin{bmatrix} 2 & 2 & -1 \\ 2 & -1 & -1 \\ -1 & -1 & -1 \end{bmatrix}$$

$$\begin{bmatrix} 2 & -1 & -1 \\ 2 & -1 & -1 \\ 2 & -1 & -1 \end{bmatrix} \begin{bmatrix} -1 & -1 & -1 \\ 2 & -1 & -1 \\ 2 & 2 & -1 \end{bmatrix} \begin{bmatrix} -1 & -1 & -1 \\ -1 & -1 & -1 \\ 2 & 2 & 2 \end{bmatrix} \begin{bmatrix} -1 & -1 & -1 \\ -1 & -1 & 2 \\ -1 & 2 & 2 \end{bmatrix}$$

Fig. 1. Compass masks

direction indices are extracted from local and extensive regions respectively. This direction information discriminates the different texture as pixel value changes.

Our approach utilizes the details from local and extensive neighborhood of reference pixel, instead of the local neighborhood that conveys primarily about the texture. Furthermore, proposed descriptor encodes the structure information into a six-bit compact code that improves the computation performance of the recognition algorithm as compared to eight bit long coded descriptors like LBP.

The proposed method can be summarized as follows:
1). It uses the max edge response directional numbers instead of intensity values, which extracting more texture information from input images.
2). It encodes the structure information in six-bit compact code by analyzing reference pixel with the local and extensive neighborhood that deliver more information with less space.
3). It utilizes the directional information that makes it more robust in case of lighting variations and white noise.

The arrangement of the paper is as follows: In Section I introduction and literature review is presented. Section-II represents RETRaIN descriptor framework. Experimental results and performance analysis is discussed in section-III. Finally, in section-IV, based on the outcomes conclusions of the work are derived.

## II. PROPOSED METHOD

The proposed descriptor uses the direction number of high magnitude edge responses to generate a six-bit pattern, which describes the salient texture features of the facial image. Consequently, this information discriminates the expressions by extracting patterns in the highly affected regions. We call these regions as the active regions. The RETRaIN descriptor primarily addresses the problem of illumination changes along with other variable factors such as age, gender and pose variations in different subject. It does so, by choosing principal response directions from local and the extended neighborhood. We generate a six-bit pattern by computing high edge responses of the input image. The high edge responses are generated by analyzing the relation between a reference pixel and it's neighborhood pixels as shown in Fig. 3. In Fig. 1., we have shown the kernels of the eight directional compass masks [22]. The process of pattern encoding, feature extraction and classification is described in the below subsections.

### A. Proposed Descriptor

The patterns of the RETRaIN is encoded by following three sequential steps. First, compass masks as shown in Fig.

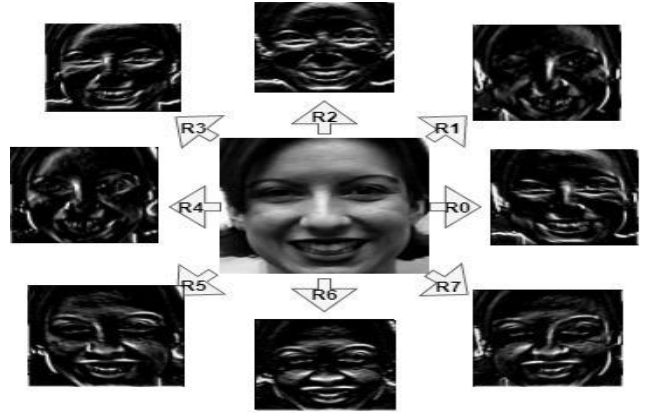

Fig. 2. Eight directional response images

1. is applied on the reference pixel of a facial image. With this operation eight edge responses are generated and further shorted to select most dominant edge response with their index position for a reference pixel. Second, the same compass masks are applied on the respective neighborhood pixels by considering directions of the mask. With this, eight edge responses are generated and further considered the dominant edge response with index positions. Finally, RETRaIN value is encoded by concatenating the binary responses of the dominant indexes generated from first two steps. The detailed representation of the proposed approaches is given in the following Eq. [1-8].

The directional compass kernels are executed on image by using Eq. (1).

$$R_\alpha = D_\alpha^m * I, 0 \leq \alpha \leq 7 \qquad (1)$$

where $I$ represents the input image, $D_\alpha^m$ is the directional compass kernel and $R_\alpha$ represents response images in different directions. Then, maximum absolute value of $R_\alpha$ is extracted from eight responses and assign as primary direction by using Eq. (2).

$$P(i,j) = \arg\max_{0 \leq \alpha \leq 7}^{\rho} \{|R_\alpha(i,j)|\} \qquad (2)$$

where $max$ is a function that returns the index value $\rho$ of maximum response set as primary direction. Further, the same directional compass mask is applied on the extended neighborhood and the edge responses in a particular direction is calculated by using below equation Eq. (3).

$$M_\alpha = D_\alpha^m * I_\alpha, 0 \leq \alpha \leq 7 \qquad (3)$$

where $I_\alpha$ represents the neighborhood space of $\alpha_{th}$ reference pixel, $M_\alpha$ is the response value in a particular direction. The index value of maximum absolute response is calculated to compute secondary direction by using Eq. (4)

$$S = \arg\max_{0 \leq \alpha \leq 7}^{\rho} \{|M_\alpha|\} \qquad (4)$$

Finally, both of the index values $0 \leq \alpha \leq 7$ are combined to generate 6-bit compact code by using Eq. (5).

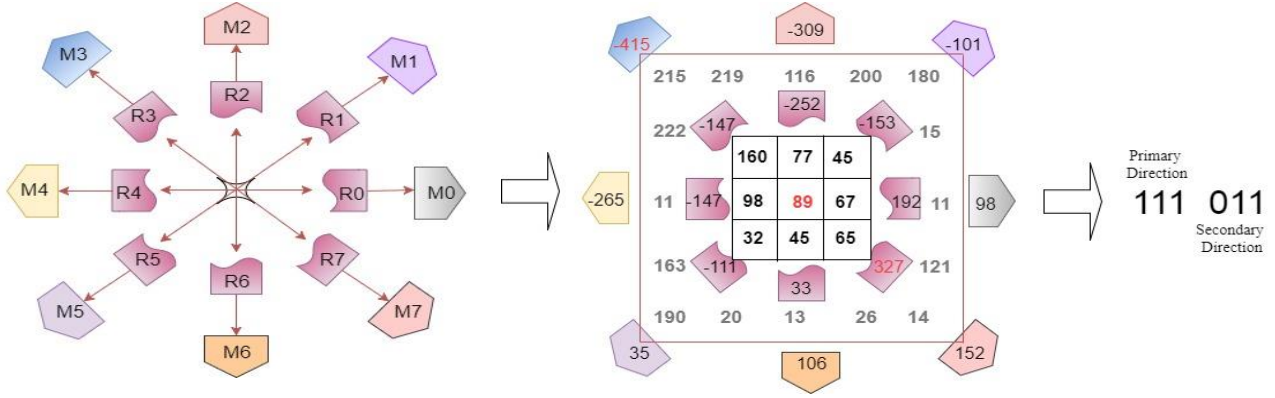

Fig. 3. Encoding scheme of proposed descriptor

$$RETRaIN(i,j) = 2^3 P(i,j) + S \qquad (5)$$

### B. Face Description

Most of the appearance-based schemes have used histograms to extract the feature vector. In this framework, the feature descriptor response is partitioned into small blocks and then histogram is created to analyze the facial features of each region. Thus, the histogram is computed using the following Eq. (6) and Eq. (7).

$$H(k) = \sum_{(i,j)\in R} \chi(RETRaIN(i,j), e), \forall e \qquad (6)$$

$$\chi(a,b) = \begin{cases} 1 & a = b \\ 0 & a \neq b \end{cases} \qquad (7)$$

where $e \in [0, 2^6 - 1]$ is pattern code range and $(i,j)$ represents coordinates in a particular region $R$.

### C. Classification

In recent time, Support Vector machine (SVM) [21] classifier has gained increased attention in machine learning. SVM is a learning technique that uses the concept of hyper decision plane to take decisions. The hyper plane separates the objects belonging to different classes. In this paper, a multiclass SVM is used to make the binary decision by using one against one approach. For this, N-fold cross-validation testing scheme is utilize to analyze the performance of RETRaIN descriptor. In N-fold cross-validation scheme, the database is casually split into N number of parts, out of these parts N-1 parts are used as training dataset and remaining as a test dataset.

## III. EXPERIMENTAL RESULTS

In this paper three standard datasets are used to evaluate the efficiency of proposed method: Extended Cohn Kanade [22], JAFFE [23], and MUG [24] database. These datasets are prepared in different environments to perform facial expression recognition algorithms. Each dataset comprises six universal emotions (Anger, Happiness, Disgust, Sadness, Surprise, and Fear) with one Neutral emotion. To validate the performance of the proposed method, we analyze obtained results with different descriptors as LBP, LDP, LDN, and LDTP which shows remarkable results. For the comparative study, we also implemented these descriptors and evaluated the performance according to our setup and datasets.

TABLE I
FACIAL EXPRESSION RECOGNITION RATE (%) OF RETRAIN AND OTHER METHODS ON CK+, JAFFE AND MUG DATABASES FOR 6 AND 7 CLASS.

| Method | CK + | | JAFFE | | MUG | |
|---|---|---|---|---|---|---|
| | 6- class | 7- class | 6- class | 7- class | 6- class | 7- class |
| LBP | 93.46 | 89.00 | 86.11 | 85.23 | 99.20 | 98.40 |
| CS- LBP | 95 | 92.55 | 88.33 | 86.68 | 99.20 | 98.57 |
| LDP | 96.25 | 92.90 | 90.55 | 86.19 | 98.80 | 98.52 |
| LDN | 94.80 | 91.68 | 83.88 | 81.42 | 98.67 | 98.57 |
| LDTP | 95.28 | 91.86 | 90.55 | 85.71 | 98.87 | 98.57 |
| RETRaIN | 97.83 | 95.16 | 91.66 | 88.09 | 99.66 | 98.57 |

### A. The Cohn Kanade Database

The Cohn Kanade (CK+) database holds 593 facial expression images of 123 university students having different age groups. From these image sequences, we used only 309 sequences with six expression states: Anger, Happy, Disgust, sad, surprise, and Fear. Neutral emotion of the same subjects is also included to make 7 class expressions. Three peak images of each emotion concerning each subject were selected to make 927 sequences of basic six expressions and 1043 sequences of seven expressions.

The recognition accuracy of the proposed approach over CK+ database is shown in Table I. With comparative encoding techniques as Local Binary Pattern (LBP) [14], CS LBP [25], Local Directional Pattern (LDP) [16], Local Directional Number (LDN) [17], and Local Directional Ternary Pattern (LDTP) [18]. Proposed method obtains better results at 97.83% for 6 expression class and 95.16% for seven expression class. To show classification rate in more details, confusion matrices are also included as shown in Table II. for both 6 class and 7 class expressions. The result shows six expressions class has more accuracy and two expressions (Disgust and Surprise) being confused with Anger and Sad expressions. In case of 7 expression class confusion rate increases as expressions get confused with the neutral expression. Thus, accuracy rate is also getting affected, and decreased compare to 6 classes.

TABLE II
CONFUSION MATRIX OF CK+ DATABASE FOR 6 CLASS FACIAL EXPRESSIONS.

|     | Ang | Dis | Fear | Hap | Sad | Sur |
|-----|-----|-----|------|-----|-----|-----|
| Ang | 26  | 0   | 0    | 0   | 0   | 0   |
| Dis | 1   | 35  | 0    | 0   | 0   | 0   |
| Fear| 0   | 0   | 15   | 0   | 0   | 0   |
| Hap | 0   | 0   | 0    | 39  | 2   | 0   |
| Sad | 0   | 0   | 0    | 0   | 17  | 0   |
| Sur | 0   | 0   | 0    | 0   | 1   | 49  |

TABLE III
CONFUSION MATRIX OF CK+ DATABASE FOR 7 CLASS FACIAL EXPRESSIONS.

|     | Neu | Ang | Dis | Fear | Hap | Sad | Sur |
|-----|-----|-----|-----|------|-----|-----|-----|
| Neu | 17  | 1   | 0   | 0    | 0   | 0   | 0   |
| Ang | 1   | 26  | 2   | 0    | 0   | 0   | 0   |
| Dis | 0   | 1   | 33  | 0    | 0   | 1   | 1   |
| Fear| 0   | 0   | 0   | 15   | 0   | 0   | 0   |
| Hap | 0   | 0   | 0   | 0    | 42  | 1   | 0   |
| Sad | 1   | 0   | 0   | 0    | 0   | 15  | 0   |
| Sur | 1   | 0   | 0   | 0    | 0   | 0   | 49  |

TABLE IV
CONFUSION MATRIX OF JAFFE DATABASE FOR 6 CLASS FACIAL EXPRESSIONS.

|     | Ang | Dis | Fear | Hap | Sad | Sur |
|-----|-----|-----|------|-----|-----|-----|
| Ang | 8   | 0   | 0    | 0   | 0   | 0   |
| Dis | 0   | 3   | 1    | 0   | 0   | 0   |
| Fear| 0   | 0   | 3    | 0   | 1   | 0   |
| Hap | 0   | 0   | 0    | 7   | 0   | 0   |
| Sad | 0   | 0   | 0    | 1   | 6   | 0   |
| Sur | 0   | 0   | 0    | 0   | 0   | 6   |

TABLE V
CONFUSION MATRIX OF 6 CLASS FACIAL EXPRESSIONS OF MUG DATABASE.

|     | Ang | Dis | Fear | Hap | Sad | Sur |
|-----|-----|-----|------|-----|-----|-----|
| Ang | 55  | 0   | 0    | 0   | 0   | 0   |
| Dis | 0   | 47  | 0    | 0   | 0   | 0   |
| Fear| 0   | 0   | 50   | 0   | 0   | 0   |
| Hap | 0   | 1   | 0    | 50  | 2   | 0   |
| Sad | 0   | 0   | 0    | 0   | 48  | 0   |
| Sur | 0   | 0   | 0    | 1   | 0   | 46  |

*B. The JAFFE Database*

Another database, The Japanese Female Facial Expression (JAFFE) [23] is used to measure the performance of the proposed scheme. This database comprises only 213 peak image sequences of ten Japanese female subjects. In this database, each subject is captured with six universal expressions (3 most expressive emotions of each subject). To perform our experiment on JAFFE, we made two datasets, 7-expression class (Anger, Happy, Disgust, Sad, Surprise, Fear, and Neutral) and 6- expression class (Anger, Happy, Disgust, Sad, Surprise, and Fear). The resultant accuracy rate of the proposed approach over JAFFE database with the existing descriptor is shown in Table I. Detailed results of JAFFE database are shown in Table. IV. Only one instance of Sad expression was classified as Happy expression.

*C. The MUG Database*

The Multimedia Understanding Group (MUG) database [24] consists of facial expression image sequences of 86 subjects (35 females and 51 male) aged from 20 to 35 years. All image sequences are labeled with universal emotion states. To test the performance of RETRaIN method 301 expression images with six basic emotions were chosen. In this setup, we use five most expressive images of each emotion so finally, we have 1505 image sequences in 6- expression class and 1760 image sequences in 7- expression class. The resultant accuracy rate of the proposed approach over JAFFE database is shown in Table I. with previously used encoding schemes. In case of MUG database Happiness and Disgust expressions shows confusion in identification as mentioned in Table V.

IV. CONCLUSION

In this paper, a new feature descriptor is proposed: RETRaIN for FER system. The RETRaIN is encoded by applying compass mask on a reference pixel and its neighborhood by considering their orientation. Further, the dominant edge responses indices are cancatenated by converting into binary pattern. Then, binary pattern is encoded into RETRaIN. From the evaluated results we can see that our proposed descriptor RETRaIN outperforms on the three standard datasets. We observe that use of extensive regions makes the proposed RETRaIN more robust and reliable descriptor as compare to other existing directional descriptors.